\documentclass[letterpaper, 10 pt, journal, twoside]{IEEEtran}
\ifCLASSINFOpdf
\else
\fi
\usepackage{graphicx}
\usepackage{subfig}
\usepackage{amsmath,amssymb,amsfonts}
\usepackage{amsthm}
\usepackage{bm} 
\newcommand{\vct}[1]{\bm{#1}}  
\newcommand{\mat}[1]{\mathbf{#1}} 

\newtheorem{myTheorem}{Theorem}[section]

\makeatletter

\newcommand{\Rmnum}[1]{\expandafter\@slowromancap\romannumeral #1@}
\makeatother

\begin{document}

\title{Interactive Force-Impedance Control}

\author{Fan Shao$^{1}$, Satoshi Endo$^{2}$, Sandra Hirche$^{2}$, {\itshape Fellow, IEEE}, and Fanny Ficuciello$^{1}$, \textit{Senior Member, IEEE}%

\thanks{$^{1}$Fan Shao and Fanny Ficuciello are with PRISMA Lab, Department of Electrical Engineering and Information Technology, University of Naples Federico II, Naples, Italy
        {\tt\footnotesize (e-mail: fan.shao@unina.it; fanny.ficuciello@unina.it).}}%
\thanks{$^{2} $Satoshi Endo and Sandra Hirche are with Chair of Information-oriented Control (ITR), Department of Electrical and Computer Engineering, Technical University of Munich, Munich, Germany
        {\tt\footnotesize (e-mail: s.endo@tum.de; hirche@tum.de).}}%
}


\maketitle

\begin{abstract}
Human collaboration with robots requires flexible role adaptation, enabling the robot to switch between an active leader and a passive follower. Effective role switching depends on accurately estimating human intentions, which is typically achieved through external force analysis, nominal robot dynamics, or data-driven approaches. However, these methods are primarily effective in contact-sparse environments. When robots under hybrid or unified force–impedance control physically interact with active humans or non-passive environments, the robotic system may lose passivity and thus compromise safety. To address this challenge, this paper proposes a unified Interactive Force-Impedance Control (IFIC) framework that adapts to interaction power flow, ensuring safe and effortless interaction in contact-rich environments. The proposed control architecture is formulated within a port-Hamiltonian framework, incorporating both interaction and task control ports, thereby guaranteeing autonomous system passivity. Experiments in both rigid and soft contact scenarios demonstrate that IFIC ensures stable collaboration under active human interaction, reduces contact impact forces and interaction force oscillations.
\end{abstract}


\IEEEpeerreviewmaketitle

\section{Introduction}

\IEEEPARstart{W}{ith} the development of lightweight collaborative robots, physical human-robot interaction (pHRI) provides operators direct and intuitive ways to manipulate robots according to their intentions. Research in this field has primarily focused on human safety, physical fatigue, and intention estimation. Among pHRI tasks, trajectory tracking and regulation (i.e., compliance control with a stationary desired pose) are the most widely studied, while relatively fewer studies have addressed direct force control. In trajectory tracking or regulation tasks, where contact is mainly induced by the human, signals such as external wrenches, joint torque, and robot dynamics can effectively convey human intentions to the robot. Human intention estimation methods include external force analysis in the time or frequency domain \cite{geravand2013human} \cite{duchaine2012stable}, nominal robot dynamics \cite{landi2017admittance},
and data-driven approaches such as model-free neural networks, model-based human impedance models, Gaussian process regression \cite{wang2023role}, and other machine learning techniques. Bayesian neural networks combined with human model learning can improve intention estimation accuracy and reduce human–robot disagreement \cite{ma2024human}. Machine learning applied to joint torque sensing has enabled intrinsic tactile perception \cite{iskandar2024intrinsic}, while multimodal intention recognition integrating vision and tactile data outperforms monomodal methods \cite{wong2023vision}. Nevertheless, these methods are generally effective in contact-sparse environments. To address the challenges of passivity and safety in contact-rich pHRI, energy tank-based methods have been proposed. A virtual energy tank supplies energy to the controller \cite{angerer2017port}, regulates impedance/admittance parameter adaptation, and attenuates the control output when needed. By defining an appropriate energy budget, the tank stores dissipated energy and compensates for non-passive behaviors that increase system energy, such as variations in inertia or stiffness in impedance control \cite{raiola2018development} \cite{ferraguti2015energy}, as well as adaptive variations of inertia and damping in admittance control to mitigate oscillatory instabilities in pHRI caused by changes in human arm stiffness \cite{ferraguti2019variable}. Furthermore, energy tanks have also been adopted in hierarchical impedance control to recover passivity lost due to null-space projections in multi-task robotic systems \cite{dietrich2016passive}. Tank energy has also been utilized as an indicator of human interaction intent, while still ensuring passivity \cite{khoramshahi2020dynamical}.

For tasks requiring simultaneous force and motion control, such as polishing and grinding, force and motion are regulated in the constrained space (C-space) and the unconstrained space (U-space), respectively. When the environment model is uncertain or unavailable, the orthogonality between these subspaces may be violated, potentially leading to loss of passivity. To address this issue, Haddadin et al. introduced the Unified Force-Impedance Control (UFIC) \cite{schindlbeck2015unified,haddadin2024unified}, which guarantees passivity via separate virtual energy tanks for force and impedance control. Specifically, the force tank absorbs potentially unbounded energy arising from stability-violating force control actions, such as contact loss or force regulation in the C-space, while the impedance tank dissipates the energy injected by large impact wrenches during collisions. Beyond ensuring passivity, the energy remaining in the tank can sustain non-passive control behaviors. Therefore, the power exchange between the energy tanks and the control system must be carefully regulated to satisfy higher-level objectives, including safety.

To alleviate this limitation of UFIC, Shahriari et al. proposed valve-based virtual energy tanks \cite{shahriari2018valve}, in which the power flow at individual ports is modulated through valve gains according to task requirements. While this extension effectively addresses controller-induced non-passive behaviors and enhances the regulation of energy exchange, UFIC still assumes a passive environment. As a result, it encounters difficulties when external interactions are non-passive. In particular, when external forces drive the robot away from the contact surface, the force controller continues to generate significant kinetic energy before the force tank reaches its lower energy bound.

Such non-passive actions can compromise safety. To cope with external non-passive interactions, existing approaches either rely on the modeling of contact and non-contact transitions based on a priori environment models \cite{salehian2018dynamical} or require additional sensing modalities. Multimodal sensing approaches that combine force/torque (F/T) measurements with additional modalities, such as electromyography (EMG) signals \cite{peternel2016towards}, tactile skins \cite{armleder2022interactive}, or vision \cite{yan2024unified}, can improve awareness of human–environment interactions, at the cost of increased hardware complexity.

To address these limitations, this paper proposes a passivity-guaranteed interaction-aware IFIC. By explicitly extracting the interaction port from the controller, virtual energy tanks are connected to both the interaction and control ports via their input–output pairs to capture and regulate the corresponding power flows. The controller output is then adaptively adjusted according to the detected non-passive interaction or control behaviors. This mechanism ensures system passivity while preserving task performance and safe whole-body physical interaction. Thanks to the power-flow–valve-controlled energy tank design, the proposed method achieves safe, responsive, and intuitive pHRI in contact-rich scenarios. The remainder of this paper is organized as follows. Section \Rmnum{2} presents the calculation of interaction power and the design of system ports. Section \Rmnum{3} details the design of the IFIC framework. Section \Rmnum{4} verifies the passivity of the proposed system. Section \Rmnum{5} describes the experimental validation, and Section \Rmnum{6} concludes the paper.

\section{Interaction Power Pair and Port Design}
\subsection{Interaction Power}
The classification of force-impedance control tasks can be based on the kinematic constraints imposed by contact surfaces \cite{suomalainen2022survey}. Under ideal conditions with orthogonal motion and force subspaces, $\dot{\vct{x}}_d^T\vct{F}_d=0$, where $\dot{\vct{x}}_d, \vct F_d\in \mathbb{R}^6$ denote the desired Cartesian velocity and wrench, respectively. This work considers the non-ideal case where the environment model is uncertain or unavailable, under which this orthogonality is violated.

Environmental or human interactions may act in both force and impedance control spaces. In most force control tasks, the force control frame aligns with the end-effector frame. Let $\mat{R}^\dagger\in \mathbb{R}^{3\times 3}$ denote the rotation matrix representing the orientation of the force control frame with respect to the world frame, and define $
\mathcal{R}^{\dagger}
=
\mathrm{diag}\big(\mathbf{R}^{\dagger},\,\mathbf{R}^{\dagger}\big)\in\mathbb{R}^{6\times 6}$. Then, $\vct F_d={\mathcal{R}}^{\dagger}\vct F^{\dagger}_d$, where $\vct F^{\dagger}_{d} \in \mathbb{R}^6$ denotes the desired wrench applied by the robot to the environment, expressed in the force frame, with the components associated with the U-space set to zero. A binary vector $\vct F^{bin}_d$ is obtained by mapping the non-zero elements of $\vct F^{\dagger}_d$ to one. The force control directional matrix in the world frame is defined as
\begin{equation}
\mathbf{D}_w\in \mathbb{R}^{6\times6}=\mathcal{R}^{\dagger}\mathrm{diag}(\vct F^{bin}_d)
\label{D_w}
\end{equation}
where $\mathrm{diag}(\vct F^{bin}_d)$ is the diagonal matrix whose diagonal entries are given by $\vct F^{bin}_d$. The C-space projection matrix is defined as the span of $\mathbf{D}_w$'s column space: $[\mathbf{D}_w]\in \mathbb{R}^{6\times 6}=\mathbf{D}_w(\mathbf{D}_w^T\mathbf{D}_w)^+\mathbf{D}_w^T$. The interaction power in the C-space is given by $P_c=\dot{\vct x}^T[\mathbf{D}_w]\vct F_{ext}$, while the interaction power in the U-space is calculated using the kernel of the force control directions to remove force control contributions: $P_u=\dot{\vct x}^T\langle\mathbf{D}_w\rangle \vct F_{ext}$, where $\langle\mathbf{D}_w\rangle=\mathbf{I}-[\mathbf{D}_w]$, and $\mathbf{I}\in\mathbb{R}^{6\times6}$ is the identity matrix.

\subsection{Port Allocation of Force and Impedance Control}
The inverse dynamic model of an $n$-DOF manipulator with revolute joints can be expressed as
\begin{equation}
\mat M(\vct q)\Ddot{\vct q}+\mat C(\vct q,\Dot{\vct q})\Dot{\vct q}+\vct g(\vct q)=\vct\tau+\mat J^T(\vct q)\vct F_{ext},
\label{inverseDynamic}
\end{equation}
where $\mat M(\vct q)\in \mathbb{R}^{n\times n}$,  $\mat C(\vct q,\dot{\vct q})\in \mathbb{R}^{n\times n}$, $\vct g(\vct q)\in \mathbb{R}^n$ are the inertia, Coriolis/centrifugal, and gravity terms, respectively; $\vct\tau\in \mathbb{R}^n$ is the joint torque; $\vct F_{ext}\in \mathbb{R}^6$ is the external wrench that the environment applies to the robot; $\mat J(\vct q)\in \mathbb{R}^{6\times n}$ is the manipulator Jacobian; and $\vct q, \Dot{\vct q}, \Ddot{\vct q} \in \mathbb{R}^n$ denote joint positions, velocities, and accelerations. Cartesian force control is realized as $\vct\tau_f=\mat J^T\vct F_f$, with
\begin{equation}
    \vct F_f=\mat K_p(\vct F_{ext}+\vct F_d)+\vct F^{i,d}_f,
\end{equation}
where
\begin{equation}
\begin{aligned}
\vct F^{i,d}_f&=\mat K_i\int_{0}^{t} (\vct F_d(\theta)+\vct F_{ext}(\theta))\, d\theta\\&+\mat K_d(\dot{\vct F}_d+\dot{\vct F}_{ext}) + \vct F_d,
 \end{aligned}
\end{equation} 
where $\mat K_p, \mat K_i, \mat K_d\in\mathbb{R}^{6\times 6}$ are gain matrices. Gains in world frame are related to force-frame gains by
\begin{equation}  
\mat K_{g} = 
\mathcal{R}^\dagger
\mat K^{\dagger}_{g}{\mathcal{R}^\dagger}^T, 
\quad g \in \{p, i, d\},
\label{K_g}
 \end{equation}
where $\mat K^{\dagger}_{p}, \mat K^{\dagger}_{i}, \mat K^{\dagger}_{d}$ are diagonal matrices, with their diagonal elements corresponding to the U-space set to zero. Considering the robot dynamics in Cartesian space:
\begin{equation}
\mat \Lambda(\vct q)\Ddot{\vct x}+\bm \mu(\vct q,\Dot{\vct q})\Dot{\vct x}+\vct F_g(\vct q)={\mat J^{+T}}\vct\tau_u+\vct F_{ext},
     \label{2.2}
\end{equation}
where $\mat\Lambda(\vct q), \bm\mu(\vct q,\Dot{\vct q}), \vct F_g(\vct q)$ are, respectively, the inertia, the Coriolis/centrifugal, and gravity in Cartesian space, and $\mat J^{+}$ is the Moore-Penrose pseudoinverse of the Jacobian \cite{khatib1995inertial}. Impedance control law in the world frame is:
\begin{equation}
\begin{aligned}
   \vct \tau_i=\mat J^T(\mat\Lambda\Ddot{\vct x}_d-\mat D_d\Dot{\tilde{\vct x}}-\mat K_s\tilde{\vct x}+\bm\mu\Dot{\vct x}_d+\vct F_g),
   \label{ImpedanceControl}
   \end{aligned}
\end{equation}
with $\dot{\vct x}_d\in \mathbb{R}^6$ the desired Cartesian velocity, $\tilde{\vct x}(t)=\vct x(t)-\vct x_d(t)$ the Cartesian tracking error, and ${\vct x}_d$ the desired Cartesian pose obtained by integrating $\dot{\vct x}_d$, $\mat D_d\in  \mathbb{R}^{6\times 6}$ and $\mat K_s\in \mathbb{R}^{6\times 6}$ the desired damping and stiffness matrices, respectively. The combined force-impedance control torque is $\vct \tau_u=\vct \tau_f+\vct \tau_i$, leading to
\begin{equation}
\mat\Lambda\Ddot{\tilde{\vct x}}+\bm\mu\dot{\tilde {\vct x}}+\mat D_d\dot{\tilde {\vct x}}+\mat K_s\tilde {\vct x}-{\vct F_f}-\vct F_{ext}=0.
\label{4.5}
\end{equation}
The system state is defined as $\tilde {\vct p}=\mat\Lambda\dot{\tilde {\vct x}}$, with storage function 
\begin{equation}
    S_{i,f}(\dot{\tilde{\vct x}},\tilde{\vct x})=\frac{1}{2}{\tilde{\vct p}}^T \mat\Lambda^{-1} {\tilde{\vct p}} +\frac{1}{2} {\tilde{\vct x}}^T \mat K_s \tilde{\vct x}. 
\end{equation}
The corresponding port-based representation of \eqref{4.5} is
\begin{equation}
\left\{
\begin{aligned}
    \begin{bmatrix}
\dot{\tilde{\vct x}} \\
\dot{\tilde{\vct p}}
\end{bmatrix}&=
\begin{bmatrix}
\mathbf{0} & \mathbf{I} \\
-\mathbf{I} & -\bm\mu-\mat D_d
\end{bmatrix}
\begin{bmatrix}
\frac{\partial S}{\partial \tilde{\vct x}} \\[5pt] 
\frac{\partial S}{\partial \tilde{\vct p}}
\end{bmatrix}+
\begin{bmatrix}
\mathbf{0} \\
\mathbf{I}
\end{bmatrix}
\underbrace{(\vct F_{ext} + \vct F_f)}_{\vct u}, \\
\vct y&=\dot{\tilde{\vct x}}.
\end{aligned}\right. \end{equation}
Here, $\vct u=\vct F_{ext} + \vct F_f$ is the system input, and $\vct y=\dot{\tilde{\vct x}}$ is the output, $\mathbf{0}, \mathbf{I}\in \mathbb{R}^{6\times6}$. With $\vct u$ as the effort and $\vct y$ as the flow, the overall system power is
\begin{equation}
\begin{aligned} P=\vct y^T\vct u&=\dot{\tilde{\vct x}}^T(\vct F_{ext}+\vct F_f)\\&=\dot{\vct x}^T\vct F_{ext}-\dot{\vct x}_d^T\vct F_{ext}+\dot{\vct x}^T\vct F_f-\dot{\vct x}^T_d\vct F_f.
\end{aligned}
\end{equation}
The system ports include the impedance control port $(-\dot{\vct x}_d, \vct F_{ext})$, the force control port $(\dot{\vct x}, \vct F_f)$, and the counteraction port $(-\dot{\vct x}_d, \vct F_f)$. Due to friction, contact loss, and impedance control counteracting force control, the control power terms $-\dot{\vct x}^T_d\vct F_{ext}$, $\dot{\vct x}^T\vct F_f$, and $-\dot{\vct x}^T_d\vct F_f$ can be positive, and passivity w.r.t the port $(\dot{\vct x}, \vct F_{ext})$ is not guaranteed. UFIC enforces passivity by augmenting the force and impedance control ports with virtual energy tanks. A force tank is associated with the force control port, whose energy rate is driven by the negative port power $-\dot{\vct x}^T\vct F_f$. Similarly, an impedance energy tank is introduced for the impedance and counteraction port, with its energy rate governed by the negative port power $\dot{\vct x}^T_d(\vct F_f+\vct F_{ext})$ and the dissipative quadratic damping term $\dot{\tilde{\vct x}}^T\mat D_d\dot{\tilde{\vct x}}$. 

The system input power of UFIC is $ P_{in}=\dot{\vct x}^T\vct F_{ext}$. Under UFIC, the time derivative of the system storage function satisfies $\dot{S}_{UFIC}\leq P_{in}$. A passive environment satisfies $\dot{S}_{env}\leq-\dot{\vct x}^T\vct F_{ext}$, which guarantees the overall passivity of system and environment, i.e., $\dot{S}_{env}+\dot{S}_{UFIC}\leq0$. However, when the environment is non-passive, the overall passivity cannot be ensured. In this case, UFIC only guarantees passivity of the robot w.r.t $(\dot{\vct x},\vct F_{ext})$, regardless of the environment dynamics. When the interaction power becomes positive ($P_{in} > 0$), the system storage function may increase, particularly when force or impedance control remains active, which can lead to excessive energy accumulation and compromise safety. Specifically, in force control space, positive interactive power may increase the robot's  potential energy, which can subsequently be converted into large kinetic energy before the force tank is depleted. In impedance control space, positive interaction power deflects the robot from the desired trajectory, thereby increasing the elastic energy stored in the impedance controller, and if $\dot{\vct x}^T_d\vct F_{ext}>0$, the impedance tank energy can further increase. After the interaction ceases, this stored energy can be released as large kinetic energy, posing safety risks.

In the IFIC framework, let $S_{sys}$ denote the storage function of the force-impedance controlled robotic system with integrated energy tanks, which satisfies $\dot{S}_{sys}\leq\dot{\vct x}^T\vct F_{ext}$. To ensure safety during interaction with non-passive environments, the overall system energy must remain bounded. Therefore, IFIC enforces autonomous passivity of the augmented system by introducing an auxiliary subsystem with storage function $S^*$ such that
\begin{equation}
\dot{S}_{IFIC}:=\dot{S}_{sys}+\dot{S}^*\leq 0. 
\end{equation}
The auxiliary system is designed to absorb non-passive interaction energy in both the C-space and the U-space, which is particularly important when either force control or impedance control is active (i.e., $\vct F_f\neq0$ or $\dot{\vct x}_d\neq0$). Using the identity $\langle\mathbf{D}_w\rangle+[\mathbf{D}_w]=\mathbf{I}$, $S^*$ is defined as
\begin{equation}
S^*=\int_{0}^{t}-k(\lambda_c\dot{\vct x}^T[\mathbf{D}_w]\vct F_{ext}+\lambda_u\dot{\vct x}^T\langle\mathbf{D}_w\rangle \vct F_{ext})d\theta,
\label{S*}
\end{equation}
where
\begin{equation}
\begin{aligned}
\lambda_{c}=
\begin{cases}
     1 & \text{if } P_c>0\land d_{f,\mathcal{I}}>0 \\
     0 & \text{else},
\end{cases} \\
\lambda_{u}=
\begin{cases}
     1 & \text{if } P_u>0\land d_{i,\mathcal{I}}>0 \\
     0 & \text{else}.
\end{cases}
\end{aligned}
\label{lambda_c}
\end{equation}
Here, $d_{f,\mathcal{I}}$ and $d_{i,\mathcal{I}}$ are defined in \eqref{d_f}, \eqref{d_i}, respectively. For $k \geq 1$, the augmented storage function $S_{IFIC}:=S_{sys}+S^*$ guarantees autonomous passivity of the IFIC system, i.e.,
\begin{equation}
\begin{aligned}
    \dot{S}^*\leq -P_{in},  \quad
    \dot{S}_{sys}+\dot{S}^*\leq 0.
\end{aligned}
\label{S.*}
\end{equation}
To implement the proposed energy regulation in the control law, the diagonal entries of $\mat K^{\dagger}_p$ associated with the U-space are set to nonzero. Since $\langle\mathbf{D}_w\rangle \vct F_d=\mathbf{0}$, the Cartesian force control law is updated as
\begin{equation}
    \begin{aligned}
        \vct F_f&=\mat K_p[\mathbf{D}_w] \vct F_{ext}+(\mat K_p[\mathbf{D}_w] \vct F_d+\vct F^{i,d}_f)\\&+\lambda_u\mat K_p\langle \mathbf{D}_w \rangle \vct F_{ext},
    \end{aligned}
    \label{ForceControl}
\end{equation}
where $\vct F_f^r := \mat K_p [\mathbf{D}_w] \vct F_d + \vct F_f^{i,d}$.
When $P_u>0$, the term $\mat K_p\langle \mathbf{D}_w \rangle \vct F_{ext}$ enforces a zero-force control behavior in the U-space.
The force control ports of the augmented system are therefore given by $(\dot{\vct x}, \mat K_p [\mathbf{D}_w] \vct F_{ext})$, $(\dot{\vct x}, \vct F_f^r)$, and $(\dot{\vct x}, \mat K_p \langle \mathbf{D}_w \rangle \vct F_{ext})$. Through these ports, the augmented system absorbs the non-passive interaction power transferred from the environment. Moreover, the interaction power can be explicitly monitored in the C-space as $\dot{\vct x}^T \mat K_p [\mathbf{D}_w] \vct F_{ext} = k_p P_c$ and in the U-space as $\dot{\vct x}^T \mat K_p \langle \mathbf{D}_w \rangle \vct F_{ext} = k_p' P_u$. The passivity condition \eqref{S.*} is discussed in Theorem~\ref{thm:passivity}.
\section{INTERACTIVE FORCE-IMPEDANCE CONTROL}
\begin{figure}[t]
\centering
    \subfloat[Force tank]{\includegraphics[scale=0.16]{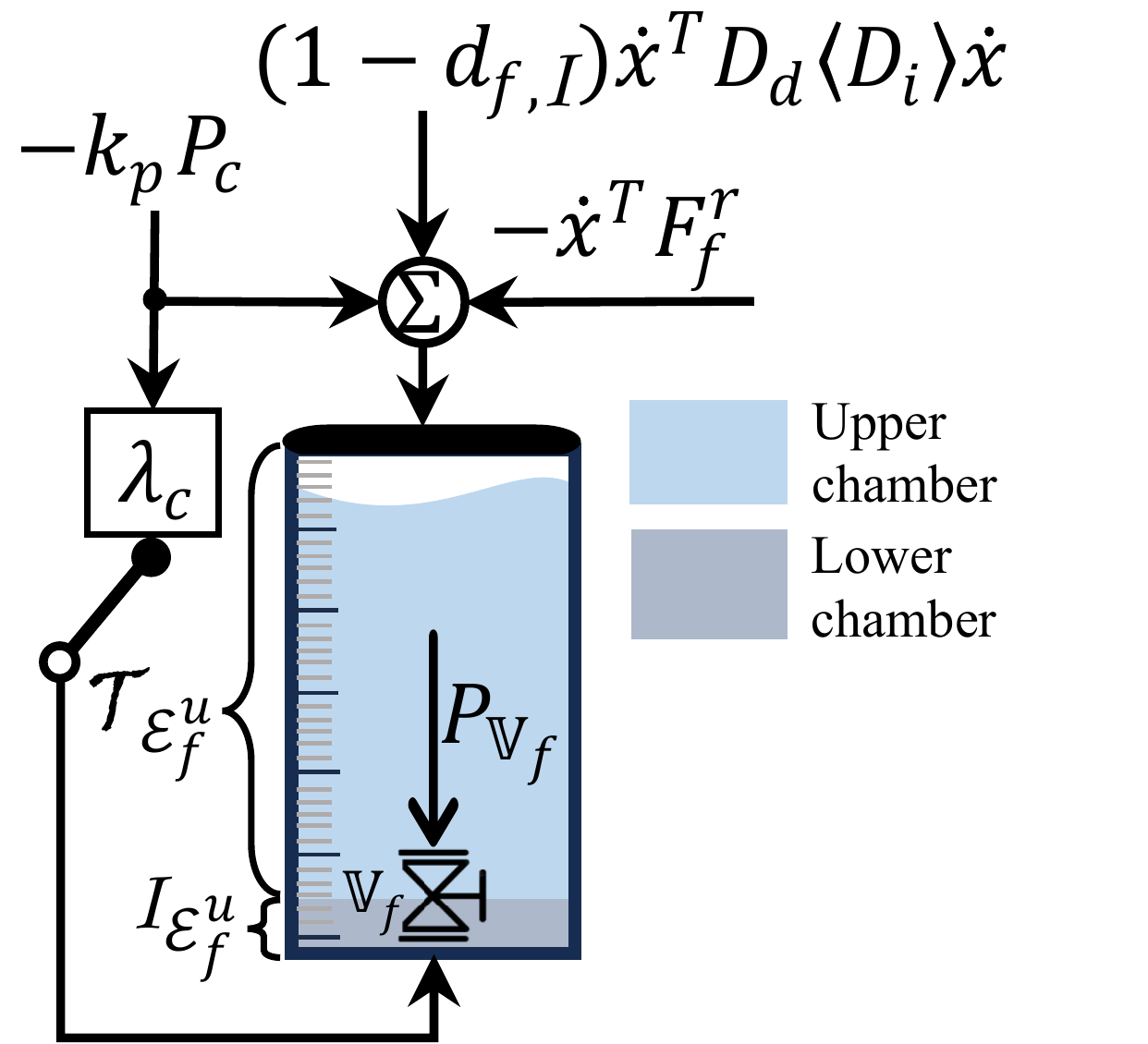}
  \label{fig:force_tank}}
	\hspace{1.7mm}
\subfloat[Impedance tank]{\includegraphics[scale=0.16]{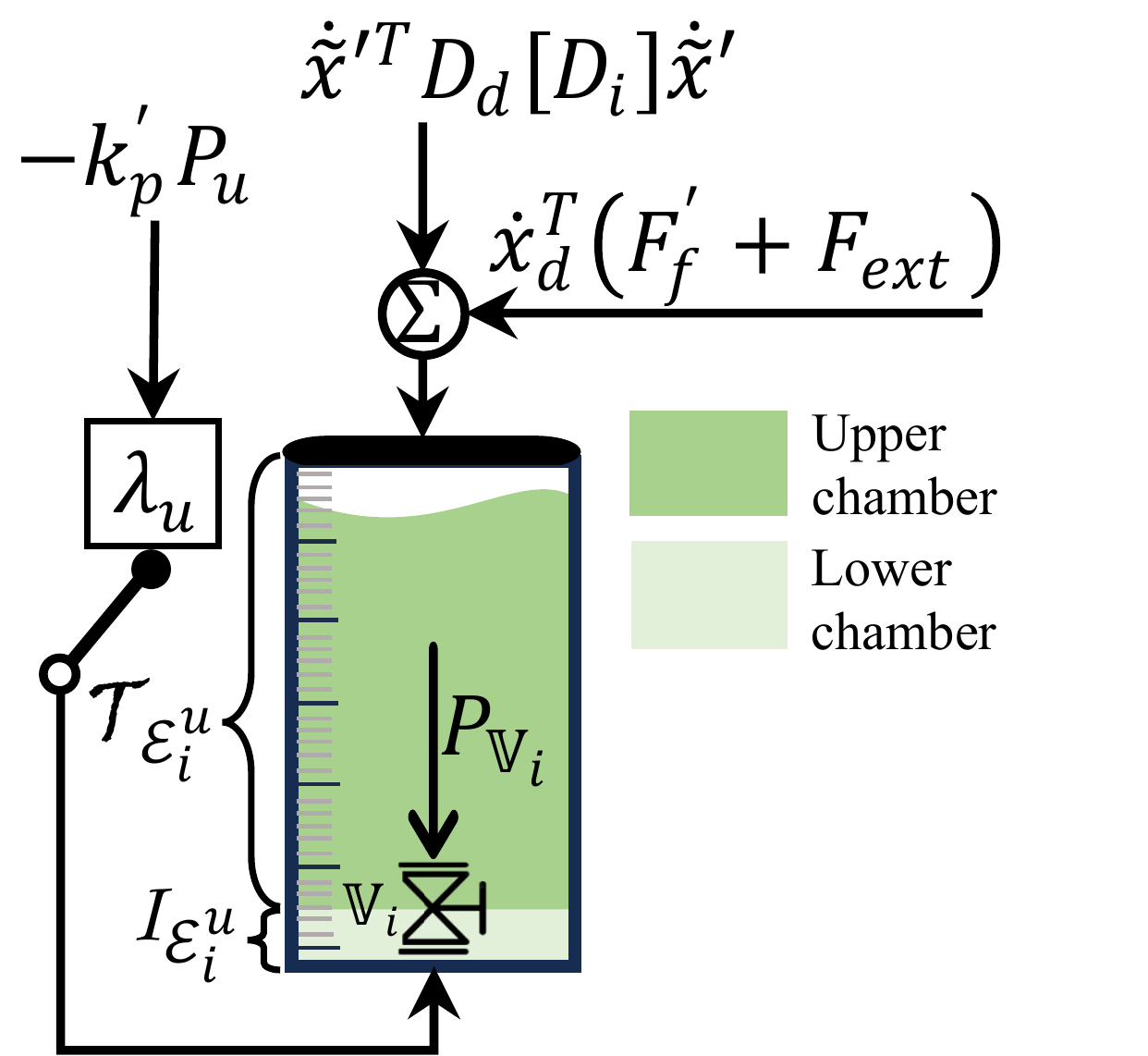}
  \label{fig:impedance_tank}}

 \caption{Valve-controlled dual-chamber tank.}
    \label{fig:tank}
\end{figure}
When a controller is augmented with virtual energy tanks, each tank handles passivity violations associated with a specific task. Multiple tasks may share a tank according to their priority levels \cite{michel2022safety}. In force–impedance control, the force and impedance tasks have equal priority, and pHRI may occur in either the C-space or U-space. Therefore, dedicated virtual energy tanks are assigned to both the force and impedance controllers.

\subsection{Interactive Force Control}
The force control tank is connected to two ports: the C-space interaction port $(\dot{\vct x}, \mat K_p[\mathbf{D}_w]\vct F_{ext})$ and the force regulation port $(\dot{\vct x}, \vct F^r_f)$, as illustrated in Fig. \ref{fig:force_tank}. It adopts a valve-controlled dual-chamber structure, where the upper (task) chamber supports the force control task and the lower (interactive) chamber absorbs non-passive interaction energy. The interactive chamber is connected to the C-space interaction port through the power gain $\lambda_c$ defined in \eqref{lambda_c}, and its energy contributes to the auxiliary storage function in \eqref{S*}. Let ${}^{\mathcal{T}}\mathcal{E}_f$ and ${}^{\mathcal{I}}\mathcal{E}_f$ denote the energies of the upper and lower chambers, respectively, with corresponding lower and upper bounds ${}^{\mathcal{S}}\mathcal{E}_{f,l}$ and ${}^{\mathcal{S}}\mathcal{E}^u_f$, $\mathcal{S}\in{\mathcal{T},\mathcal{I}}$. When either chamber energy approaches its lower bound, the force control is gradually deactivated. 

Energy transfer from the task to the interactive chamber is regulated by a power flow valve $\mathbb{V}_f$ at a rate $P_{\mathbb{V}_f}\geq 0$, defined as
\begin{equation}
\begin{aligned}
    P_{\mathbb{V}_f}=
    \begin{cases}
       P^*_{c} & \text{if } P_c\geq P^*_{c}\\
       \dfrac{{}^{\mathcal{I}}\mathcal{E}^u_f}{t_f} & \text{else}.
    \end{cases}
\end{aligned}
\end{equation}
Here, $P^*_{c}$ defines the interaction power threshold. A small upper bound ${}^{\mathcal{I}}\mathcal{E}^u_f$ (e.g., 0.1 $\mathrm{J}$) ensures rapid depletion of the interactive chamber when the C-space interaction power exceeds $P^*_{c}$, enabling fast response to non-passive interactions. When $P_c < P^*_{c}$, the interactive chamber is recharged at a constant rate determined by the loading time $t_f$. The threshold $P^*_c$ trades off sensitivity to non-passive interactions and robustness of force regulation: smaller values increase responsiveness to non-passive interaction power, whereas larger values preserve stable force regulation under disturbances. In this work, a small threshold ($P_c^*=0.03$ $\mathrm{W}$) is selected to evaluate the system’s responsiveness to external interactions.

The upper chamber energy ${}^{\mathcal{T}}\mathcal{E}_f=0.5z^2_f$ with state $z_f$ is defined as 
\begin{equation}
\begin{aligned}
    \dot{z}_f=\frac{1}{z_f}(\mathcal{P}_f+(1-d_{f,\mathcal{I}})\dot{\vct x}^T\mat D_d\langle \mathbf{D}_i \rangle \dot{\vct x}-P_{\mathbb{V}_f}),
\end{aligned}
    \label{forceTank}
\end{equation}
where $\mathbf{D}_i \in \mathbb{R}^{6\times6}$ is the impedance control directional matrix, $\langle \mathbf{D}_i \rangle=\mathbf{I}-[\mathbf{D}_i]$, the upper tank chamber input is
\begin{equation}
\begin{aligned}
\mathcal{P}_f&=-d_{f,\mathcal{T}}d_{f,\mathcal{I}}(\dot{\vct x}^T\vct F^r_f+\dot{\vct x}^T\mat K_p[\mathbf{D}_w]\vct F_{ext}),
\end{aligned}
\label{forceTankInput}
\end{equation}
where the damping terms $d_{f,\mathcal{T}}, d_{f,\mathcal{I}}\in[0,1]$ are computed from the tank energy:
\begin{equation}
\underset{\mathcal{S}\in\{\mathcal{T},\mathcal{I}\}}{d_{f,\mathcal{S}}} = 
\begin{cases} 
1\hspace{1.4cm} \text{if }\hspace{0.2cm}^{\mathcal{S}}\mathcal{E}_f\geq  {}^{\mathcal{S}}\mathcal{E}_{f,l}+{}^{\mathcal{S}}\delta_{f,h} \\[2pt]
0 \hspace{1.4cm} \text{else if } \hspace{0.2cm} {}^{\mathcal{S}}\mathcal{E}_{f}<{}^{\mathcal{S}}\mathcal{E}_{f,l}+{}^{\mathcal{S}}\delta_{f,s} \\[2pt]
\cos\left((1-\mathcal{A}^p\right)\frac{\pi}{2}) \hspace{0.74cm} \text{else},
\end{cases}
\label{d_f}
\end{equation}
with
\begin{equation}
    \begin{aligned}
\mathcal{A}=\frac{{}^{\mathcal{S}}\mathcal{E}_{f}-{}^{\mathcal{S}}\mathcal{E}_{f,l}-{}^{\mathcal{S}}\delta_{f,s}}{{}^{\mathcal{S}}\delta_{f,h} - {}^{\mathcal{S}}\delta_{f,s}}, \hspace{0.2cm}
p=\begin{cases}
    10\hspace{0.74cm} \text{if } P_c>P^*_c\\[2pt]
    1\hspace{0.5cm} \text{else}.
\end{cases}
\end{aligned}
\end{equation}
When the energy approaches ${}^{\mathcal{S}}\mathcal{E}_{f,l}$, $d_{f,\mathcal{S}}$ smoothly transitions from 1 to 0 over the interval $[{}^{\mathcal{S}}\mathcal{E}_{f,l}+{}^{\mathcal{S}}\delta_{f,s},{}^{\mathcal{S}}\mathcal{E}_{f,l}+{}^{\mathcal{S}}\delta_{f,h}]$. The parameter $p$ governs the decay rate of $d_{f,\mathcal{S}}$: larger $p$ values correspond to faster decay. This design ensures that the force control becomes inactive immediately when a non-passive interaction is detected ($P_c>P^*_c$) and gradually recovers once the interaction becomes passive. The interactive chamber energy $ {}^\mathcal{I}\mathcal{E}_f(t)=0.5{z^*_f}^2$, with state $z^*_f$, is governed by
\begin{equation}
\begin{aligned}
    \dot{z}^*_f=\frac{1}{z^*_f}(
        P_{\mathbb{V}_f}-\lambda_{c} \dot{\vct x}^T\mat K_p[\mathbf{D}_w]\vct F_{ext}).
\end{aligned}
\label{z*f}
\end{equation}
The coupling between the task and interactive chambers will not affect task performance, since the dissipated power of the quadratic damping terms is recycled to recharge the task chamber during interaction.
Note that \eqref{d_f} and \eqref{lambda_c} prevent energy singularities (${}^{\mathcal{S}}\mathcal{E}_{f}<0$) but does not enforce the upper bound ${}^{\mathcal{S}}\mathcal{E}^u_{f}$. We therefore directly set ${}^{\mathcal{S}}\mathcal{E}_{f}={}^{\mathcal{S}}\mathcal{E}^u_{f}$ when ${}^{\mathcal{S}}\mathcal{E}_{f}>{}^{\mathcal{S}}\mathcal{E}^u_{f}$. This choice avoids introducing additional parameters and reduces complexity. If $P_c<0$, the augmented system in the C-space forms a power-preserving
Dirac structure; otherwise, the injected interaction energy is dissipated by the auxiliary subsystem.
Finally, the force control output is modified by
\begin{equation}
\vct F'_f=d_{f,\mathcal{T}}d_{f,\mathcal{I}}\vct F_f.
    \label{F'_f}
\end{equation}
$d_{f,\mathcal{I}}$ modulates the damping term in C-space, where the desired velocity is set to zero in \eqref{forceTank}, to ensure power consistency, the impedance control can be rewritten as
\begin{equation}
\begin{aligned}
   \vct \tau_i
   &=\mat J^T(\mat \Lambda\Ddot{\vct x}_d-((1-d_{f,\mathcal{I}}) \mat D_d\langle \mathbf{D}_i\rangle\dot{\vct x}+  \mat D_d[\mathbf{D}_i]\dot{\tilde{\vct x}})\\&-\mat K_s\tilde{\vct x}+\bm\mu\Dot{\vct x}_d+\vct F_g).
   \label{ImpedanceControl2}
   \end{aligned}
\end{equation}
In this case, impedance control remains active in C-space when force control is deactivated.

\subsection{Interactive Impedance Control}
The impedance tank adopts the same dual-chamber design as the force tank and is connected to two ports: the U-space interaction port $(\dot{\vct x},\mat K_p\langle \mathbf{D}_w\rangle \vct F_{ext})$ and the impedance control port $(-\dot{\vct x}_d,\vct F'_f+\vct F_{ext})$ (Fig. \ref{fig:impedance_tank}). Let ${}^{\mathcal{T}}\mathcal{E}_i$ and ${}^{\mathcal{I}}\mathcal{E}_i$ denote the energies of the upper (task) and lower (interactive) chambers, respectively, with upper bounds ${}^{\mathcal{T}}\mathcal{E}_i^u$ and ${}^{\mathcal{I}}\mathcal{E}_i^u$. The task chamber supports the impedance control task, whereas the interactive chamber absorbs non-passive interaction energy in the U-space. In the absence of external interactions, frictional effects yield $P_u<0$. When $P_u>0$, the interactive chamber is connected to port $(\dot{\vct x},\mat K_p\langle \mathbf{D}_w\rangle \vct F_{ext})$. The power flow rate is set to $P_{\mathbb{V}_i} = 0.01$ $\mathrm{W}$ for $P_u > 0$, and to $P_{\mathbb{V}_i} = {}^{\mathcal{I}}\mathcal{E}^u_i / t_{i}$ for $P_u < 0$.
The upper chamber energy ${}^{\mathcal{T}}\mathcal{E}_i=0.5z^2_i$, with state $z_i$, evolves according to
\begin{equation}
\begin{aligned}
\dot{z}_i&=\frac{1}{z_i}(d_{i,\mathcal{T}}d_{i,\mathcal{I}}\dot{\vct x}^T_d(\vct F'_f+\vct F_{ext})+\dot{\tilde{\vct x}}'^T\mat D_d[\mathbf{D}_i]\dot{\tilde{\vct x}}'-P_{\mathbb{V}_i}),
\end{aligned}
\label{impedance_tank}
\end{equation}
where $\dot{\tilde{\vct x}}'=\dot{\vct x}-d_{i,\mathcal{T}}d_{i,\mathcal{I}}\dot{\vct x}_d$. $d_{i,\mathcal{T}}$ and $d_{i,\mathcal{I}}$ are computed as
\begin{equation}
\underset{\mathcal{S}\in\{\mathcal{T},\mathcal{I}\}}{d_{i,\mathcal{S}}} = 
\begin{cases} 
1\hspace{1.4cm} \text{if }\hspace{0.2cm}^{\mathcal{S}}\mathcal{E}_i\geq  {}^{\mathcal{S}}\mathcal{E}_{i,l}+{}^{\mathcal{S}}\delta_{i,h} \\[2pt]
0 \hspace{1.4cm} \text{else if } \hspace{0.2cm} {}^{\mathcal{S}}\mathcal{E}_{i}<{}^{\mathcal{S}}\mathcal{E}_{i,l}+{}^{\mathcal{S}}\delta_{i,s} \\[2pt]
\cos\left((1-\mathcal{A}^p\right)\frac{\pi}{2}) \hspace{0.74cm} \text{else},
\end{cases}
\label{d_i}
\end{equation}
with
\begin{equation}
    \begin{aligned}
\mathcal{A}=\frac{{}^{\mathcal{S}}\mathcal{E}_{i}-{}^{\mathcal{S}}\mathcal{E}_{i,l}-{}^{\mathcal{S}}\delta_{i,s}}{{}^{\mathcal{S}}\delta_{i,h} - {}^{\mathcal{S}}\delta_{i,s}}, \hspace{0.04cm}
p=\begin{cases}
    10\hspace{0.74cm} \text{if } P_u>0\\[2pt]
    1\hspace{0.5cm} \text{else},
\end{cases}
\end{aligned}
\end{equation}
where ${}^{\mathcal{S}}\mathcal{E}_{i,l}$ denotes the lower energy limit and ${}^\mathcal{S}\delta_{i,h}>{}^\mathcal{S}\delta_{i,s}$. The interactive chamber energy is defined as ${}^\mathcal{I}\mathcal{E}_i=0.5{z^*_i}^2$, with dynamics
\begin{equation}
\dot{z}^*_i=
\frac{1}{z^*_i}(P_{\mathbb{V}_i}-2\lambda_{u} \dot{\vct x}^T\mat K_p\langle\mathbf{D}_w\rangle \vct F_{ext}). 
\label{z*i}
\end{equation}
Accordingly, the desired Cartesian velocity is modified as:
\begin{equation}
\dot{\vct x}'_d=d_{i,\mathcal{T}}d_{i,\mathcal{I}}\dot{\vct x}_d.
\label{v'd}
\end{equation}

\section{PASSIVITY PRESERVING ANALYSIS}
After augmenting the system with dual-chamber energy tanks, $\vct F'_f=d_{f,\mathcal{T}}d_{f,\mathcal{I}}\vct F_f$, $\tilde {\vct p}'=\mat \Lambda\dot{\tilde{\vct x}}'$. The storage function is
\begin{equation}
    S_{IFIC}=\frac{1}{2}{\tilde{\vct p}}'^T \mat\Lambda^{-1} {\tilde{\vct p}'} +\frac{1}{2} {\tilde{\vct x}}'^T \mat K_s \tilde{\vct x}'+\frac{1}{2}(z^2_f+{z^*_f}^2+z^2_i+{z^*_i}^2). 
    \label{V}
\end{equation}
Let $\vct X$ be the system state and $\vct \nabla \vct S_{IFIC}$ denote the gradient of $S_{IFIC}$. For brevity, let $S := S_{IFIC}$.
\begin{equation}
\begin{aligned}
    \dot{\vct X}:&=[\dot{\tilde{\vct x}}'^T,\dot{\tilde{\vct p}}'^T,\dot{z}_f, \dot{z}^*_f, \dot{z}_i, \dot{z}^*_i]^T, \\
\vct \nabla \vct S_{IFIC} :&= 
\left[\frac{\partial S}{\partial \tilde{\vct x}'}^T, 
\frac{\partial S}{\partial \tilde{\vct p}'}^T, 
\frac{\partial S}{\partial z_f}, \frac{\partial S}{\partial z^*_f},
\frac{\partial S}{\partial z_i},\frac{\partial S}{\partial z^*_i}\right]^T.
\end{aligned}
\end{equation}
The input vector $\vct U:=[\vct u^T\ \vct u^T_f\ {\vct u^*_f}^T \ \vct u^T_i \ {\vct u^*_i}^T]^T$, with
\begin{equation}
\begin{alignedat}{2}
\vct u & = \vct F_{ext}+\vct F'_f, \quad & \vct u_f & = -\mat K_p[\mathbf{D}_w]\vct F_{ext}-\vct F^r_f,\\
\vct u^*_f & = -\lambda_c\mat K_p[\mathbf{D}_w]\vct F_{ext}, \quad & \vct u_i & = \vct F'_f+\vct F_{ext},\\
\vct u^*_i & = -2\lambda_u\mat K_p\langle\mathbf{D}_w\rangle \vct F_{ext}.
\end{alignedat}
\end{equation}
The overall system can thus be expressed in port form 
\begin{equation}
\left\{
\begin{aligned}
\dot{\vct X}
&=
\mathrm{diag}\!\left(
\mat G,
\mat H\right)\vct \nabla \vct S_{IFIC}
+ \mat B\vct U,\\[3pt]
\vct Y&=\mat B^T\vct \nabla \vct S_{IFIC},
\end{aligned}
\right.
\label{eq:compact-system}
\end{equation}
where $\mat B\in \mathbb{R}^{16\times30}$, $\mat G\in\mathbb{R}^{12\times12}$, $\mat H\in\mathbb{R}^{4\times4}$
\begin{equation}
\mat B=
\begin{bmatrix}
\mathbf{0} & & & & \\[-1pt]
\mathbf{I} & & & & \\[-2pt]
& \vct b_f & & & \\[-1pt]
& & \vct b_f^* & & \\[-1pt]
& & & \vct b_i & \\[-1pt]
& & & & \vct b_i^*
\end{bmatrix},\ 
\left\{
\begin{aligned}
\vct b_f   &= \tfrac{d_{f,\mathcal T}d_{f,\mathcal I}}{z_f}\,\dot{\vct x}^T,\\
\vct b_f^* &= \tfrac{1}{z_f^*}\,\dot{\vct x}^T,\\
\vct b_i   &= \tfrac{d_{i,\mathcal T}d_{i,\mathcal I}}{z_i}\,\dot{\vct x}_d^T,\\
\vct b_i^* &= \tfrac{1}{z_i^*}\,\dot{\vct x}^T,
\end{aligned}
\right.
\end{equation}

\begin{equation}
\begin{aligned}
\mat G=
\begin{bmatrix}
\mathbf{0} & \mathbf{I} \\
-\mathbf{I} & -\bm\mu-\mat D_d
\end{bmatrix},\ \mat H=\mathrm{diag}(h_f,h^*_f,h_i,h^*_i),
\end{aligned}
\end{equation}
where $h_f=z^{-2}_f((1-d_{f,\mathcal{I}})\dot{\vct x}^T\mat D_d\langle \mathbf{D}_i \rangle \dot{\vct x}-P_{\mathbb{V}_f})$, $h^*_f={z^*_f}^{-2}P_{\mathbb{V}_f}$, $h_i=z^{-2}_i(\dot{\tilde{\vct x}}'^T\mat D_d[\mathbf{D}_i ] \dot{\tilde{\vct x}}'-P_{\mathbb{V}_i})$, $h^*_i={z^*_i}^{-2}P_{\mathbb{V}_i}$.
The instantaneous power of the entire system becomes
\begin{equation}
    \begin{aligned} 
\vct Y^T\vct U&=\dot{\tilde{\vct x}}'^T\vct u+d_{f,\mathcal{T}}d_{f,\mathcal{I}}\dot{\vct x}^T\vct u_f+\dot{\vct x}^T\vct u^*_f+\dot{\vct x}'^T_d\vct u_i+\dot{\vct x}^T\vct u^*_i\\
&=\dot{\vct x}^T(\vct F_{ext}+\vct F'_f)-\dot{\vct x}_d'^T(\vct F_{ext}+\vct F'_f)\\&-d_{f,\mathcal{T}}d_{f,\mathcal{I}}\dot{\vct x}^T(\mat K_p[\mathbf{D}_w]\vct F_{ext}+\vct F^r_f)\\&-\lambda_c\dot{\vct x}^T\mat K_p[\mathbf{D}_w]\vct F_{ext}+\dot{\vct x}'^T_d(\vct F_{ext}+\vct F'_f)\\&-2\lambda_u\dot{\vct x}^T\mat K_p\langle\mathbf{D}_w\rangle \vct F_{ext}\\
&=\dot{\vct x}^T\vct F_{ext}-\lambda_c\dot{\vct x}^T\mat K_p[\mathbf{D}_w]\vct F_{ext}\\&-\lambda_u\dot{\vct x}^T\mat K_p\langle\mathbf{D}_w\rangle \vct F_{ext}.
    \end{aligned}
    \label{YU}
\end{equation}

\begin{myTheorem}
\label{thm:passivity}
If the force gain in the force control frame satisfies
$\mat K^{\dagger}_p = k \mathbf{I}$ with $k \geq 1$, then $\vct Y^T \vct U \leq 0$ when either force control or impedance control is active.
\end{myTheorem}
\begin{proof}
Since power is coordinate-independent, all terms are expressed in the force control frame for simplicity.
Let $\mathbf{D}^\dagger_w = \mathrm{diag}(\vct F^{bin}_d)$,
$[\mathbf{D}^\dagger_w] + \langle \mathbf{D}^\dagger_w \rangle = \mathbf{I}$, $\vct F^\dagger := \vct F^{\dagger}_{ext}$.
The interaction powers in the C-space and U-space are given by
\[
P_c = \dot{\vct x}^{\dagger T}[\mathbf{D}^\dagger_w]\vct F^\dagger, \quad
P_u = \dot{\vct x}^{\dagger T}\langle \mathbf{D}^\dagger_w \rangle \vct F^\dagger,
\]
with $\dot{\vct x}^{\dagger T}\vct F^\dagger = P_c + P_u$.
Substituting into \eqref{YU} yields
\[
\vct Y^T \vct U
= (1 - k\lambda_c) P_c + (1 - k\lambda_u) P_u.
\]
According to the definitions of $\lambda_c$ and $\lambda_u$, when $k \geq 1$, both coefficients are non-positive whenever the corresponding interaction power is positive. Therefore, $\vct Y^T \vct U \leq 0$ holds in all cases.
\end{proof}
 
By combining \eqref{4.5}, \eqref{ForceControl}, \eqref{forceTank}, \eqref{z*f}, \eqref{F'_f}, \eqref{ImpedanceControl2} \eqref{impedance_tank}, \eqref{z*i}, \eqref{v'd} and \eqref{V}, the time derivative of the IFIC storage function is obtained as
\begin{equation}
\begin{aligned}
    \dot{S}_{IFIC}&=\dot{\vct x}^T\vct F_{ext}-\dot{\tilde{\vct x}}'^T\mat D_d([\mathbf{D}_i]+\langle \mathbf{D}_i \rangle)\dot{\tilde{\vct x}}'\\&+\dot{\vct x}^T\vct F'_f-\dot{\vct x}_d'^T(\vct F'_f+\vct F_{ext})\\
    &-(\lambda_c+d_{f,\mathcal{T}}d_{f,\mathcal{I}})\dot{\vct x}^T\mat K_p[\mathbf{D}_w]\vct F_{ext}\\
    &-d_{f,\mathcal{T}}d_{f,\mathcal{I}}\dot{\vct x}^T\vct F^r_f+(1-d_{f,\mathcal{I}})\dot{\vct x}^T\mat D_d\langle \mathbf{D}_i \rangle \dot{\vct x}\\
&+d_{i,\mathcal{T}}d_{i,\mathcal{I}}\dot{\vct x}^T_d(\vct F'_f+\vct F_{ext})+\dot{\tilde{\vct x}}'^T\mat D_d[\mathbf{D}_i]\dot{\tilde{\vct x}}'\\&-2\lambda_u\dot{\vct x}^T\mat K_p\langle\mathbf{D}_w\rangle \vct F_{ext} \\
&=\underbrace{\dot{\vct x}^T\vct F_{ext}-d_{f,\mathcal{I}}\dot{\vct x}^T\mat D_d\langle \mathbf{D}_i \rangle \dot{\vct x}}_\text{$\dot{S}_{sys}$}\\&\underbrace{-\lambda_c\dot{\vct x}^T\mat K_p[\mathbf{D}_w]\vct F_{ext}-\lambda_u\dot{\vct x}^T\mat K_p\langle\mathbf{D}_w\rangle \vct F_{ext}}_\text{$\dot{S}^*$}.
    \end{aligned}
    \label{dotV}
\end{equation}
It follows directly that $\dot{S}_{sys}\leq \dot{\vct x}^T\vct F_{ext}$. Under the condition $\mat K^{\dagger}_p=k\mathbf{I}$ with $k\geq1$, Theorem~\ref{thm:passivity} guarantees that $\dot{S}^*\leq -\dot{\vct x}^T\vct F_{ext}$.
Therefore, $\dot{S}_{IFIC}:=\dot{S}_{sys}+\dot{S}^*\leq0$, which proves that the proposed IFIC framework preserves passivity. As a result, safe pHRI is ensured when either force control or impedance control is active.

\begin{table*}[t]
\vspace{3mm}
\renewcommand{\arraystretch}{1.3} 
\setlength{\tabcolsep}{5pt} 
 \caption{Experimental parameters}
    \centering
    \begin{tabular}{@{}lllllllllllllll@{}}
\hline
\rule{0pt}{2.5ex}%
  &$f^\dagger_{d,z}$ & $P_{\mathbb{V}_f}$ & $P_{\mathbb{V}_i}$& ${}^{\mathcal{I}}\mathcal{E}^u_f$& ${}^{\mathcal{I}}\mathcal{E}^u_i$& ${}^{\mathcal{T}}\mathcal{E}^u_f$& ${}^{\mathcal{T}}\mathcal{E}^u_i$ &$t_{f}$&$t_{i}$& $\mat K^\dagger_p$& $\mat K^\dagger_i$& $\mat K^\dagger_d$ &$k_{s,t}$, $k_{s,r}$ & $d_{d,t}$, $d_{d,r}$\\
  \hline
 &(N) & (W) & (W)& (J)& (J)& (J)& (J) & (s)&(s) & &[1/s]& (s) &[N/m], [Nm/rad] & [N·s/m], [Nm·s/rad]\\

Exp. 1 &10  & 0.03&0.01&0.1&0.1&1&1&2&2&2$\mathbf{I}_{6\times6}$&2$\mathbf{I}_{6\times6}$&0.02$\mathbf{I}_{6\times6}$&800, 25&300, 3\\

Exp. 2 & 10 & 0.03&0.01&0.1&0.1&1&1&2&2&$\mathbf{I}_{6\times6}$&0.5$\mathbf{I}_{6\times6}$&0.01$\mathbf{I}_{6\times6}$&800, 10&200, 1  \\

Exp. 3 & 3 & 0.05&0.01&0.1&0.1&2&1&2&2&2$\mathbf{I}_{6\times6}$&2$\mathbf{I}_{6\times6}$&0.01$\mathbf{I}_{6\times6}$&1500, 25&300, 3 \\

Exp. 4 & 3 & 0.05&0.01&0.1&0.1&2&1&2&2&2$\mathbf{I}_{6\times6}$&2$\mathbf{I}_{6\times6}$&0.01$\mathbf{I}_{6\times6}$&1500, 25&300, 3 \\
\hline
\end{tabular}
    \label{tab:parameter}
\end{table*}
\section{EXPERIMENTS}
\begin{figure}[t]
    \centering
    \includegraphics[width=0.95\linewidth]{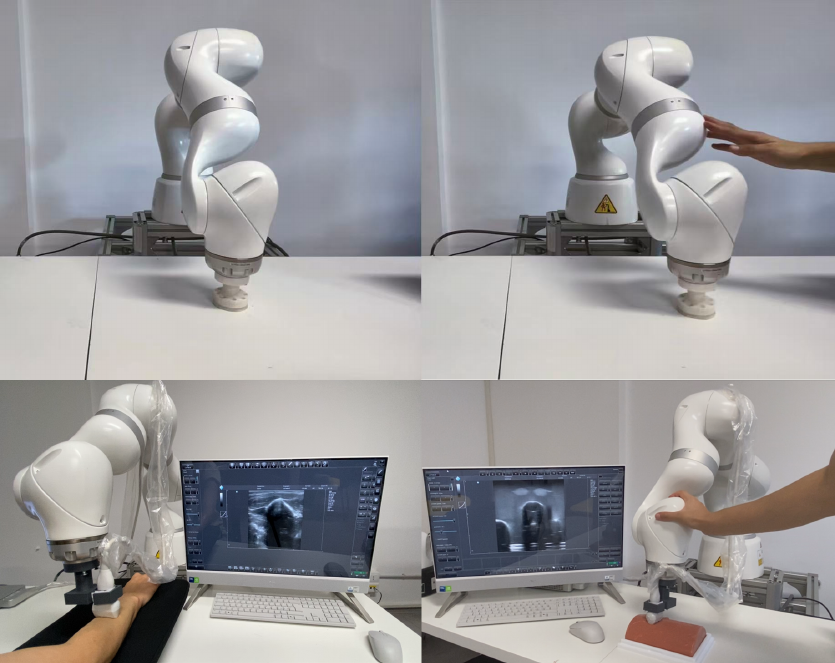}
    \caption{Experimental setup. Top: interactive table-wiping task. Bottom: interactive ultrasound scanning task on human arm and soft-tissue phantom.}
    \label{fig:setup}
\end{figure}

Two pHRI scenarios were considered: robotic table wiping and robot-assisted ultrasound scanning on a soft phantom and a human arm, as illustrated in Fig.~\ref{fig:setup}. The experiments were conducted using a KUKA LBR Med R820 robot with a 1~kHz Fast Robot Interface (FRI), operating in joint torque control mode. Joint torque sensors were used to estimate the external wrench $\vct F_{ext}$, enabling full-body physical interaction.

The experiments evaluate the performance of the proposed IFIC under active interactions from non-passive environments. For details on task energy budget computation and task fulfillment, readers are referred to the UFIC framework~\cite{haddadin2024unified}. IFIC is compared with UFIC using identical force and impedance controllers, identical controller parameters, and the same task energy budgets, as summarized in TABLE~\ref{tab:parameter}.

In the table wiping task, the desired contact force was set to $f^\dagger_{d,z}=10~\mathrm{N}$ along the vertical $z$-direction, while the desired velocity in the $xy$-plane followed sinusoidal trajectories, resulting in a periodic reciprocating motion. In robot-assisted ultrasound scanning, the target force was set to $f^\dagger_{d,z}=3~\mathrm{N}$ along the $z$-direction. Four experiments were conducted. We first demonstrate IFIC in real table wiping. We then perform a controlled Gazebo simulation of the same task to systematically sweep the injected interaction energy and quantify impact forces under repeatable conditions. Finally, we compare IFIC and
UFIC in interactive ultrasound scanning on a soft phantom and on a human arm, respectively. 

\begin{figure}[b]
    \centering
    \includegraphics{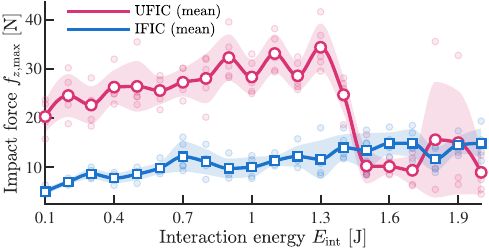}
    \caption{Comparison of peak impact forces under different interaction energy levels in Gazebo simulation.}
    \label{fig:simulation}
\end{figure}

\begin{figure*}[t]
  \centering
\subfloat[Interactive table-wiping task within U-space.]
{\includegraphics{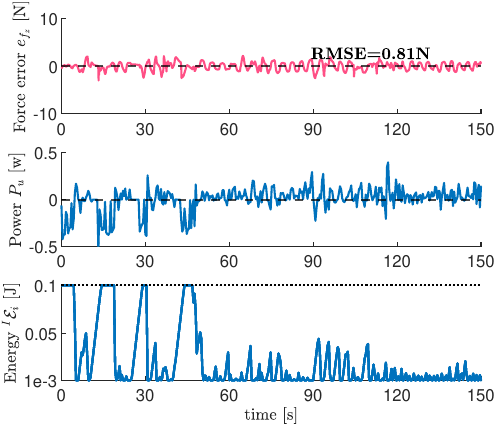}
  \label{fig:e1-uspace}}
  \hfill
   \subfloat[Interactive table-wiping task within C-space.]{ \includegraphics{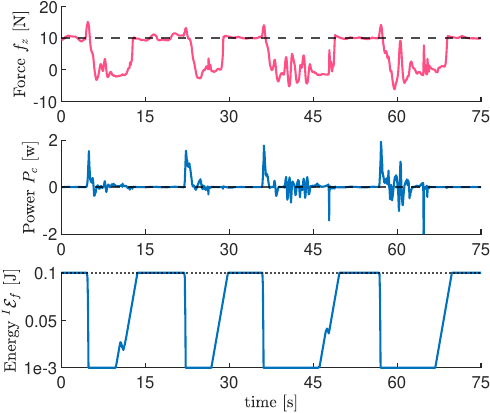}
    \label{fig:e1-cspace}}
    \caption{Evaluation of IFIC in an interactive table-wiping task.}
    \label{fig:e1}
\end{figure*}

The U-space interaction results of IFIC in the table-wiping task are shown in Fig.~\ref{fig:e1-uspace}. During the initial stage ($t<50~\mathrm{s}$), the operator applied impulsive disturbances along the impedance control direction. The interactive tank energy ${}^\mathcal{I}\mathcal{E}_i$ decreased promptly under positive interaction power $P_u$, which reduced the effective impedance and made the robot compliant during external interaction. After the disturbance, impedance control was recovered within $t_{i}=2~\mathrm{s}$. During the remaining stage ($50\leq t\leq150~\mathrm{s}$), the operator provided continuous guidance in the U-space. Throughout the task, the desired contact force was tracked with an RMSE of $0.81~\mathrm{N}$. The C-space interaction results are shown in Fig.~\ref{fig:e1-cspace}. When intended or unintended interactions drove the robot out of the workspace, the force controller was immediately deactivated upon detecting positive $P_c$. Once the operator released the robot, the interactive tank energy ${}^{\mathcal{I}}\mathcal{E}_f$ recharged, allowing force control to resume. The small impact force along the $z$-axis indicates that IFIC enables safe physical interaction.

In the second experiment, we conducted a controlled simulation study to
systematically explore the safety limits of UFIC and IFIC under
repeatable interaction conditions. The table-wiping task was simulated
in Gazebo using a KUKA LBR Med R820 model with identified dynamic
parameters from the real robot. To induce contact loss, external forces
were applied along the $z$-axis to lift the end-effector away from the
table and then released to trigger re-contact.
We quantify the impact severity by the peak contact force
$f_{z,\max}$ during the first 100~ms after contact initiation. The
injected interaction energy $E_{int}$ was swept from $0.1~\mathrm{J}$ to
$2.0~\mathrm{J}$ in steps of $0.1~\mathrm{J}$. For each energy level, five
independent trials were performed for both UFIC and IFIC. To ensure a
consistent excitation across conditions, the external force magnitude
was fixed at $30~\mathrm{N}$ for all energy levels, and the force was
removed once $E_{int}$ reached the prescribed target value.
As shown in Fig.~\ref{fig:simulation}, hollow circles and squares denote
the mean $f_{z,\max}$ for UFIC and IFIC, respectively, while lighter
markers indicate individual trials. The shaded envelopes represent
$\mu\pm\sigma$, where $\mu$ and $\sigma$ are the mean and standard
deviation. IFIC consistently yields lower impact forces than UFIC for
$E_{int}<1.4~\mathrm{J}$, indicating safer transient contact behavior.
For larger injected energies ($E_{int}\geq 1.4~\mathrm{J}$), UFIC may show
reduced impacts because the force tank becomes fully depleted after the
robot is lifted sufficiently high, which deactivates force control;
however, UFIC then fails to recover force regulation after the
interaction. In contrast, IFIC maintains bounded impact forces while
allowing force control to be smoothly re-established after contact.

\begin{figure}[t]
\includegraphics[trim=0 3.8cm 0 0, clip]{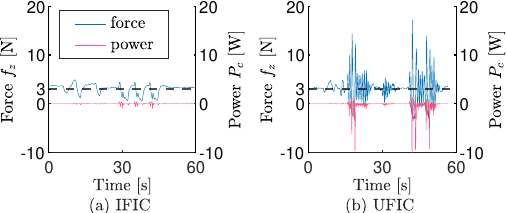}
    \caption{Comparison of IFIC and UFIC in an interactive ultrasound scanning task on a soft-tissue phantom.}
    \label{fig:e4}
\end{figure}

In the third experiment (Fig.~\ref{fig:e4}), IFIC and UFIC were compared during interactive ultrasound scanning on a soft-tissue phantom. The operator applied perturbations to the robot body to drive it away from the phantom while recording the $z$-axis contact force and $P_c$. Under these disturbances, IFIC maintained stable interaction and avoided collisions with the phantom, whereas UFIC exhibited instability with repeated bouncing. 

\begin{figure}[t]
\includegraphics[trim=0 3.8cm 0 0, clip]{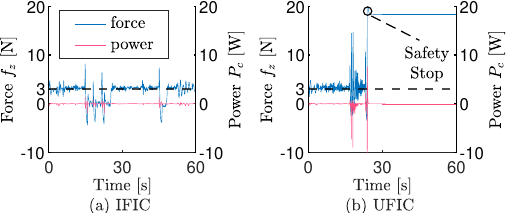}
    \caption{Comparison of IFIC and UFIC in an interactive ultrasound scanning task on human arm.}
    \label{fig:e5}
\end{figure}

In the fourth experiment (Fig. \ref{fig:e5}), interactive ultrasound scanning was performed on a human arm. During probe scanning, sudden upward arm motions occasionally caused temporary probe detachment. IFIC responded safely to such disturbances, whereas UFIC caused sudden robot jumps. In UFIC, the contact velocity occasionally exceeded the configured KUKA Med safety limit ($0.3~\mathrm{m/s}$), which triggered a safety stop.

In summary, when the robot is driven to lose contact or rebounds after re-contact, the interaction can inject positive energy into the system and thereby break passivity. In such cases, the interaction may increase the robot’s gravitational potential energy. Similarly, when the robot deviates from the desired trajectory, the elastic energy stored in the impedance controller can increase. After the interaction ceases, these stored energies may be released as kinetic energy, which may compromise safety for both the environment and the human operator. To ensure autonomous passivity when either force control or motion control remains active, IFIC augments the C-space and U-space interaction ports with power-valve-controlled energy tanks. In the C-space, IFIC absorbs non-passive interaction energy and temporarily deactivates force control to avoid contact-loss-induced instability. In the U-space, IFIC absorbs non-passive interaction energy and temporarily scales the desired velocity to zero to prevent elastic energy accumulation. In contrast, UFIC does not guarantee autonomous passivity under non-passive environments. During contact loss in the C-space, the force tank may not be sufficiently depleted, so the force controller can remain active during contact re-establishment.
In our real-robot experiments (with task energy budgets of
$1$--$2~\mathrm{J}$), the end-effector was typically lifted only a small
distance, since the active force controller resists separation from the
surface. Consequently, the force-control power
$\dot{\vct x}^{T}\vct F_f$ was often insufficient (in magnitude or
duration) to reliably drain the UFIC force tank to its lower bound. In the ultrasound scanning experiments, soft-contact rebound may even lead to intermittent phases of negative
force-control power, which can replenish the tank and prolong force
controller activity. Moreover, in the U-space, human guidance opposing
$\dot{\vct x}_d$ may be insufficient to reduce the impedance-tank energy,
while guidance aligned with the desired motion direction can increase it. By contrast, in simulation we can deliberately prescribe larger injected
interaction energies (e.g., $E_{int}>1~\mathrm{J}$), which lift the robot
higher and create larger post-release velocities. In this case, the
force-control power can become sufficiently positive to drain the UFIC
force tank, leading to deactivation of force control (as observed for
high $E_{int}$ in Fig.~\ref{fig:simulation}). This highlights the
scenario-dependence of UFIC tank depletion and further motivates IFIC,
which deactivates force control based on non-passive interaction power
rather than waiting for the tank to be depleted.

\section{CONCLUSIONS}
This paper improves safety in force–impedance control. For passive environments, the controller–tank interconnection preserves the standard power-preserving Dirac structure. For non-passive interactions, we deliberately switch from a power-preserving interconnection to a dissipative one, so that
the injected energy is absorbed and autonomous passivity is enforced ($\dot{S}_{IFIC}\leq0$). Moreover, the programmable inter-chamber power valve provides a natural
handle to adjust disturbance sensitivity across scenarios, and could be
combined with environment-awareness modules (e.g., multimodal sensing)
to enable adaptive tuning. Nevertheless, although IFIC can effectively reduce the robot kinetic energy following non-passive interactions and mitigate unstable behaviors compared with conventional methods, it does not explicitly constrain the kinetic energy. Consequently, some undesired transient responses may still arise after the interaction, including large impact forces or bouncing in the C-space and high-acceleration motion in the U-space. As shown in Fig.~\ref{fig:simulation}, under high interaction energy ($E_{int}>1~\mathrm{J}$), the robot can be lifted far away from the contact surface; when contact is re-established, the kinetic energy increases significantly, leading to larger impact forces. Future work will therefore focus on introducing explicit kinetic energy constraints and guaranteeing contact stability under a wide range of contact conditions.

\ifCLASSOPTIONcaptionsoff
  \newpage
\fi

\bibliographystyle{IEEEtran}
\bibliography{main}

\end{document}